\documentclass[conference]{IEEEtran}
\IEEEoverridecommandlockouts
\usepackage{cite}
\usepackage{amsmath,amssymb,amsfonts}
\usepackage{algorithmic}
\usepackage{graphicx}
\usepackage{textcomp}
\usepackage{xcolor}

\usepackage{algorithm}
\usepackage{algorithmic}
\usepackage{CJK}
\usepackage{times}
\usepackage{latexsym}
\usepackage{multirow}
\usepackage{booktabs}
\usepackage{algorithm}
\usepackage{algorithmic}
\usepackage{amsmath}
\usepackage{amssymb}
\usepackage{CJK}
\usepackage{graphicx}
\usepackage{subfigure}
\usepackage{bibentry}

\def\BibTeX{{\rm B\kern-.05em{\sc i\kern-.025em b}\kern-.08em
    T\kern-.1667em\lower.7ex\hbox{E}\kern-.125emX}}
\begin{document}

\title{Bilingual Terminology Extraction from Comparable E-Commerce Corpora
}



\author{
\IEEEauthorblockN{
\textbf{Hao Jia$^1$$^{\ast}$\thanks{$^{\ast}$ Hao Jia and Shuqin Gu make the equal contributions.},
Shuqin Gu$^2$$^{\ast}$,
Yuqi Zhang$^2$,
Xiangyu Duan$^1$$^{\ddagger}$\thanks{$^{\ddagger}$ Xiangyu Duan is the corresponding author.}
}
}

\IEEEauthorblockA{
\textit{$^1$Institute of Artificial Intelligence, School of Computer Science and Technology,
Soochow University} \\
\textit{$^2$Machine Intelligence Technology Lab, Alibaba DAMO Academy} \\
hjia@stu.suda.edu.cn,
shuqingu@tju.edu.cn,
chenwei.zyq@alibaba-inc.com,
xiangyuduan@suda.edu.cn
}
}



\maketitle

\begin{abstract}
Bilingual terminologies are important machine translation resources in the field of e-commerce, which are usually either manually translated or automatically extracted from parallel data. The human translation is costly and e-commerce parallel corpora is very scarce. However, the comparable data in different languages in the same commodity field is abundant. 
In this paper, we propose a novel framework of extracting e-commercial bilingual terminologies from comparable data. Benefiting from the cross-lingual pre-training in e-commerce, our framework can make full use of the deep semantic relationship between source-side terminology and target-side sentence to extract corresponding target terminology.
Experimental results on various language pairs show that our approaches achieve significantly better performance than various strong baselines.

\end{abstract}

\begin{IEEEkeywords}
bilingual terminology extraction, e-commerce domain, cross-lingual pre-training
\end{IEEEkeywords}

\section{Introduction}
In recent years, many work has indicated that user-provided or domain-specific bilingual terminologies can enhance the accuracy and consistency of machine translation in specific domain\cite{hasler2018neural,hokamp2017lexically,post2018fast,arthur2016incorporating}.
Especially in the field of e-commerce\cite{song2019code}, due to the diversity of product description, many terminologies\footnote{In this paper, e-commercial terminology refers to the key phrase that can describe product attributes, such as product name, product brand, product material, and product style, etc. } have their specific translations in specific product category, which makes it rather difficult for vanilla machine translation models to express their correct meanings. Moreover, the wrong translation of terminologies will lead to the decline of the whole sentence translation quality.
As shown in Table \ref{tbl-example}, the terminology \begin{CJK}{UTF8}{gbsn}
``\textit{大款}" (a terminology describing that the size of the cloth is big) is mistranslated into ``big money" by Google Translate\footnote{https://translate.google.com/. We present the translation results on January 10th, 2022}, and its correct translation in the sentence should be ``big size". 
Obviously, when consumers browse this product on the e-commerce website, it will be misleading to consumers because of the wrong translation of the terminology describing the product attributes. Therefore, the correct translation of terminologies is of great significance to improving the translation quality in the field of e-commerce.

\begin{table}
\caption{Example of the wrong terminology translation leading to the misunderstanding of e-commercial sentence.}\label{tbl-example}
\scriptsize
\centering
\begin{tabular}{l|l}
\toprule[1.2pt]
Source Sentence & 看来我只能买这种\textbf{大款}的 \\
\hline \hline
Google Translate & It seems that i can only buy this kind of \textbf{\emph{big money}} \\
Correct Translation & It seems that i can only buy this kind of \textbf{\emph{big size}} \\
\bottomrule[1.2pt]
\end{tabular}
\end{table}

\end{CJK}

The acquisition of bilingual terminology pairs is either manual translation or automatically extracted from parallel data \cite{gaussier1998flow,gaussier2000term,chambers2000automatic,le2008mutual,haque2014bilingual}. Manual translation is a reliable way, but it is very time-consuming and expensive.
The latter methods are either rule-based or statistical-based, using the linguistic feature, statistical feature or a hybrid of them. They rely on linguistic analysis tools, such as POS taggers, which may not be available for low-resource languages or domains. 

The above automatic extraction methods are not suitable for e-commerce because of the lack of parallel e-commerce data. However, there are large-scale monolingual corpora covering different languages on popular e-commerce platforms. In such data, there are many potential terminology pairs, which are translations of each other. How to discover these bilingual terminologies is a big challenge in the  e-commerce domain.

In this paper, we propose a new task, which is to discover bilingual terminologies from comparable data. The detailed description of constructing comparable data is presented in Section \ref{data-construction}. Here, we focus on the e-commerce field. 
Given a terminology phrase in source language and a sentence in target language, the task is to 1) distinguish whether the target sentence contains the corresponding target translation of the source terminology, and 2) extract the corresponding target terminology from the target sentence if it contains. 

To tackle this task, we propose an effective two-stage e-commercial bilingual terminology extraction framework. In the first stage,
we fine-tune a cross-lingual pre-training model with a large number of e-commercial corpus consisting of different languages.
In the second stage, we extract the target terminology from the target sentence by utilizing the extraction model initialized by cross-lingual pre-trained language models. 

The main contributions of this paper can be summarized as follows:
\begin{itemize}
\item A new task of extracting bilingual terminologies of e-commerce from comparable data is proposed. In addition, we construct the corresponding comparable data in e-commerce domain.

\item For the first time, the task of extracting bilingual terminologies of e-commerce from comparable data is formalized by using cross-lingual pre-training model and extraction framework.
\item We conduct experiments mainly on three different e-commercial categories, namely clothes category, toys category and outdoors category in Chinese-to-English and English-to-French language pairs. 
Experimental results prove the effectiveness of the method. We hope our work would inspire new paradigms for bilingual terminology extraction.
\end{itemize}

\section{Related Work}

\subsection{Cross-Lingual Word Embeddings for Bilingual Lexicon Induction}
Following the success of word embeddings \cite{mikolov2013distributed} trained on monolingual data, a large proportion of research aimed at mapping word embeddings into a common space for multiple languages \cite{zhang2017adversarial,conneau2017word,lample2017unsupervised,lazaridou2015hubness}, which were implemented by optimizing a linear transformation matrix. Based on these efforts, \cite{artetxe2018emnlp} proposed the extension of skip-gram to learn n-gram embeddings and mapped them to a shared space to obtain cross-lingual n-gram embeddings. However, these n-gram embeddings are based on the co-occurrence frequency.

\subsection{Bilingual Terminology Extraction from Parallel or Comparable Corpora}
Several influential approaches \cite{le2008mutual,kupiec1993algorithm,haque2014bilingual,gaussier2000term} have been proposed to extract bilingual terminology from parallel corpus, which mainly rely on the linguistic feature, statistical feature or the hybrid of them. \cite{kupiec1993algorithm} proposed an algorithm, which adopted English and French text taggers to associate noun phrases in the aligned English-to-French parallel corpus. The taggers provided part-of-speech categories which were used by finite-state recognizers to extract simple noun phrases for both languages. \cite{lefever2009language} proposed a sub-sentential alignment terminology extraction module that links linguistically motivated phrases in parallel texts. 
In addition, \cite{dejean2002bilingual} proposed how to optimally combine different models derived from different resources for bilingual terminology extraction from comparable corpora.
However, unlike our methods, these feature-driven (statistics or lingualistics) methods are usually not language-independent, and lack semantic information.


\subsection{Supervised Word Alignment Based on Cross-language Span Prediction}
Researchers defined the alignment as an object for indicating the corresponding words in a parallel text \cite{och2003systematic,koehn2007moses}.
Recently, \cite{nagata2020supervised} formalized the supervised word alignment method as a cross-language span prediction problem similar to the SQuAD-style question answering task \cite{rajpurkar2016squad}. Specifically, given a target sentence as the context and a source word as a question, the word alignment system predicted a
translation of the source word as the answer, which
was a span in the target sentence. 

Their idea is a little similar to our bilingual terminology extraction task based on cross-lingual pre-training model. However, in our method, in order to enhance the semantic representation, we utilize the e-commercial bilingual terminology pair and source term with corresponding target sentence pair to fine-tune the cross-lingual pre-training model. While \cite{nagata2020supervised} just utilized the multilingual BERT \cite{Devlin2019BERTPO} as the semantic feature extractor. 

\begin{figure}[t]
\flushleft
\centering
\includegraphics[width=0.48\textwidth]{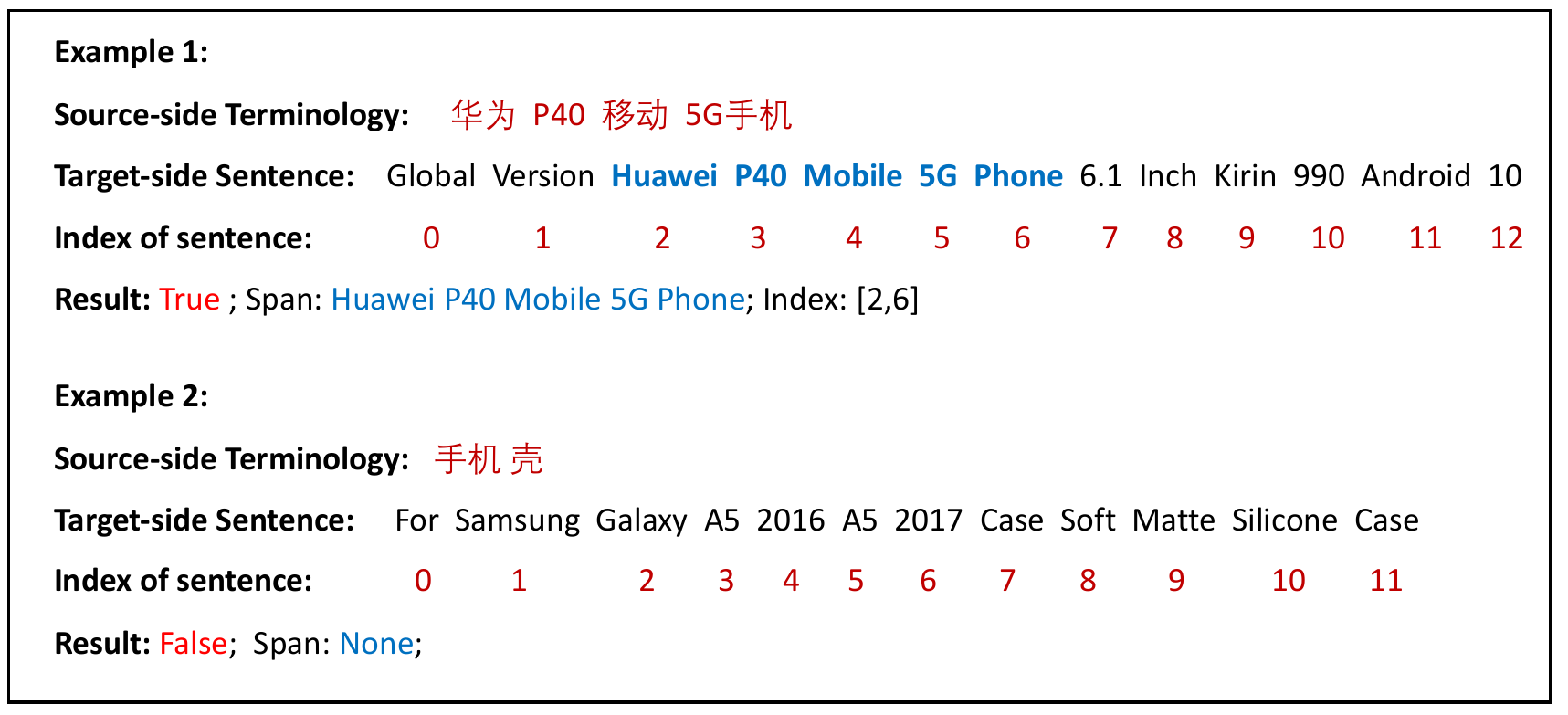}
\caption{Examples of our proposed task. Given a source-side terminology and a target-side sentence, we aim to distinguish whether the target-side sentence contains the translation of source-side terminology, and extract the corresponding translation if contains. Example 1 is the positive case, and Example 2 is the negative case.}
\label{fig:example}
\end{figure}

\section{Proposed Task and Solutions}
To our knowledge, we are the first to propose the task of e-commercial bilingual terminology extraction based on comparable corpus, independent on any parallel sentences. The definition of our proposed task and the solutions we proposed will be presented in the following.

\begin{figure*}[t]
\flushleft
\centering
\subfigure[]{
\includegraphics[width=0.48\textwidth]{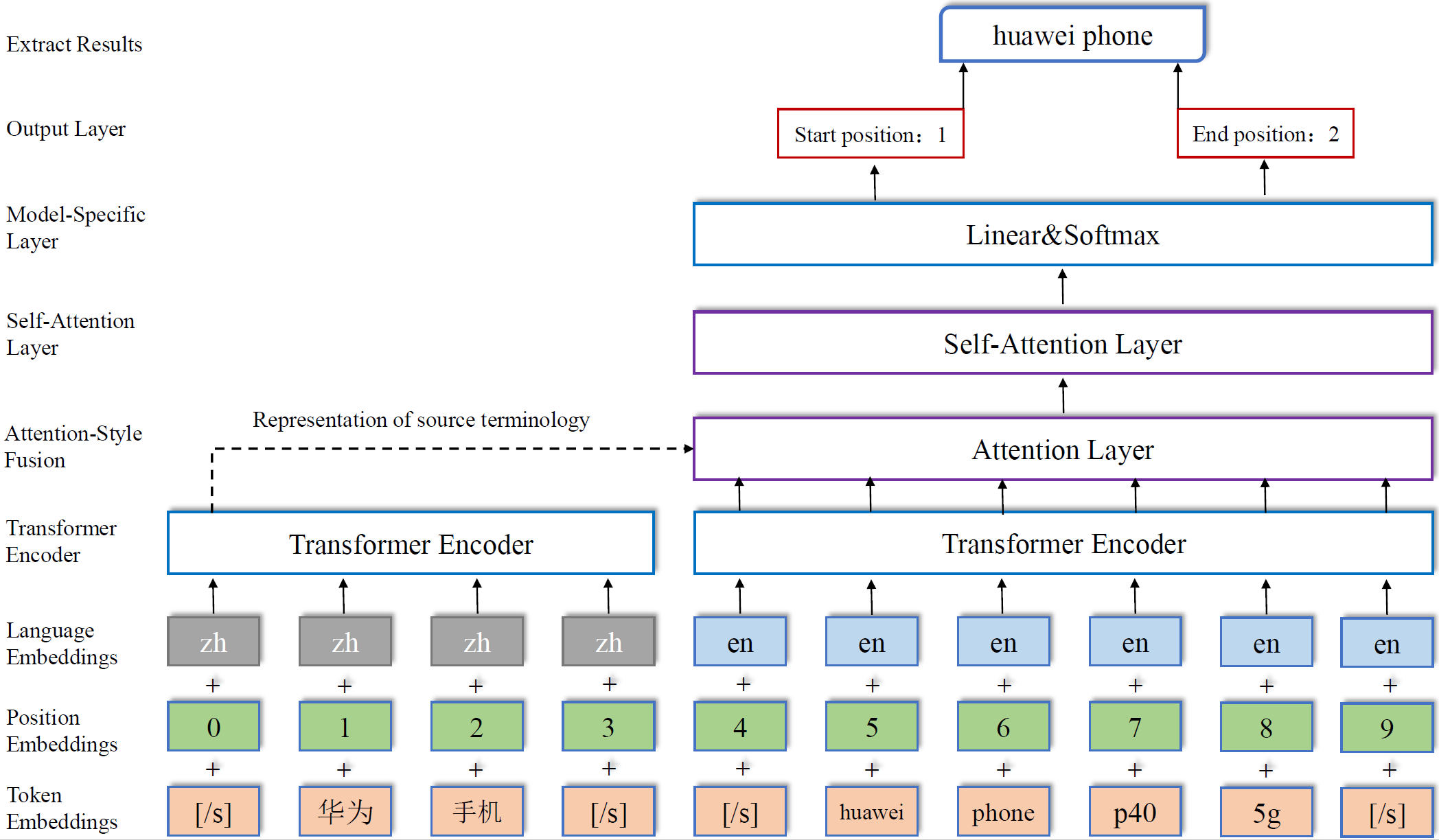}
}
\subfigure[]{
\includegraphics[width=0.48\textwidth]{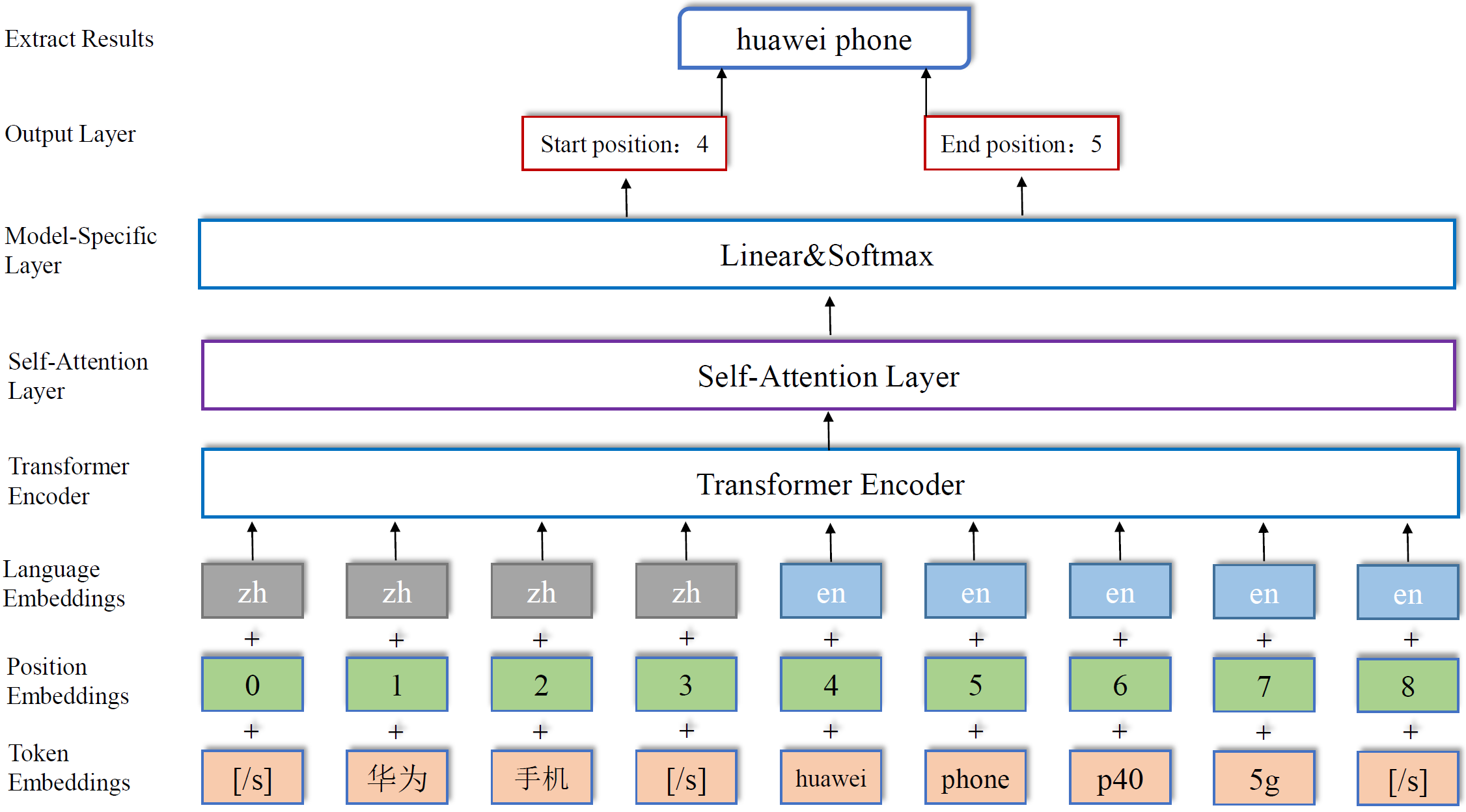}
}
\caption{The overview of our proposed methods, (a) Extractor\_Attn and (b) Extractor\_Concat. }
\label{fig:model}
\end{figure*}

\subsection{Task Definition}
Our proposed task of e-commercial bilingual terminology extraction aims to extract the potential bilingual terminologies in massive e-commercial non-parallel corpus, which can be described as commonly specialized phrases with lengths of 2-5 grams. 

In detail, given a terminology in source language and an e-commercial sentence in target language, we aim to distinguish whether the target sentence contains the corresponding target translation of the source terminology. If contains, we expect the system to find the position of the target terminology correctly. For example, as shown in Figure \ref{fig:example}, the terminology in source language is 
\begin{CJK}{UTF8}{gbsn}
``华为 P40 移动 5G手机"
\end{CJK}
, and the sentence in target language is ``Global Version Huawei P40 Mobile  5G  Phone 6.1 Inch Kirin 990 Android 10", the task is to predict the span of the potential terminology spans in the target sentence, i.e., ``Huawei P40 Mobile 5G Phone", if not exists, return None.

\subsection{Our Approach}
To tackle this task, we propose the two-stage e-commercial bilingual terminology extraction framework. In the first stage, we employ a large number of e-commercial corpus consists of different languages to perform cross-lingual pre-training. In the second stage, we extract the target terminology from the target sentence by utilizing Extractor\_Attn or Extractor\_Concat (illustrated in Figure \ref{fig:model}) initialized by cross-lingual pre-trained language models in e-commerce.

\vspace{5 pt}
\noindent  \textbf{Cross-lingual Pre-training in E-Commerce}
\vspace{5 pt}

The cross-lingual language model pre-training (XLM) \cite{conneau2019cross} method contains Masked Language Model (MLM) objective and Translation Language Model (TLM) objective, and has demonstrated its effectiveness on tasks such as XNLI cross-lingual classification and unsupervised machine translation.
MLM is conducted over large monolingual corpora by randomly masking words, and training to predict them as a Cloze task \cite{taylor1953cloze}. Since MLM is only dependent on monolingual corpora, TLM is designed to utilize parallel data to drive better alignment between source and target language representations, by the means of concatenating parallel sentences into a whole sentence and randomly masking words in both the source and target side.

Inspired by these, we adopt MLM objective to perform cross-lingual pre-training on monolingual e-commercial corpus, which is the mixture of monolingual product titles in e-commerce domain from different languages.
Besides, to gain better alignment between source and target language representations, 
we further propose to conduct TLM over the training sets of bilingual terminology pairs and the pairs of source terminology and target sentence in e-commerce.

\vspace{5 pt}
\noindent  \textbf{Target-side Terminology Extraction}
\vspace{5 pt}

Figure \ref{fig:model} generally illustrates our proposed framework. Given an e-commercial source terminology $S_{term}$ consisting of $m$ tokens $\{s_1, s_2 ... s_m\}$, we need to extract its corresponding translation span $T_{term}$ from the target sentence $T$ containing $n$ tokens $\{t_1, t_2 ... t_n\}$ . We use the Transformer\cite{Vaswani2017AttentionIA} encoder initialized by cross-lingual pre-trained models in e-commerce as the backbone to fully extract the deep semantic relationship between the source-side terminology and target-side sentence, so that our framework could correctly distinguish or even extract the target-side terminology. We propose two methods Extractor\_Attn and Extractor\_Concat to proceed the extraction of representation.

\textbf{\emph{Extractor\_Attn}:} As illustrated in Figure \ref{fig:model}(a), $S_{term}=\{[/s], s_1, s_2 ... s_m, [/s]\}$ and $T=\{[/s], t_1, t_2 ... t_n, [/s]\}$, consisting of the adding sum of language embedding, position embedding and token embedding of corresponding tokens, are fed into the Transformer encoder respectively to get the representation matrix $H_{src\_term} \in \mathbb{R}^{(m+2) \times d}$ and $H_{tgt\_stc} \in \mathbb{R}^{(n+2) \times d}$. Then we obtain the final representation matrix $H$ by doing representational fusion between source-side terminology and target-side sentence as follows:
\begin{equation}
    \begin{split}
        F(H_{tgt\_stc})=MultiHead(Q=H_{tgt\_stc},\\
        K=H_{src\_term},V=H_{src\_term})
    \end{split}
\end{equation}
\begin{equation}
    \begin{split}
        H_1=FFN(LayerNorm(H_{tgt\_stc}+F(H_{tgt\_stc})))
    \end{split}
\end{equation}
\begin{equation}
    \begin{split}
        F(H_1)=MultiHead(Q=H_1, K=H_1, V=H_1)
    \end{split}
\end{equation}
\begin{equation}
    \begin{split}
        H=FFN(LayerNorm(H_1+F(H_1)))
    \end{split}
\end{equation}
where $MultiHead$, $LayerNorm$, $FFN$ are basic components of the Transformer model. By the attention-style fusion and self-attention computation in this way, the model can utilize the representation of source-side terminology as weight to attend the most related span of target-side sentence. Note that the encoders of source-side terminology and target-side sentence share the same parameters.

\textbf{\emph{Extractor\_Concat}:} As illustrated in Figure \ref{fig:model}(b), the input sequence consists of $S_{term}$ concatenated with $T$, i.e., $\{[/s], s_1, s_2 ... s_m, [/s], t_{1}, t_{2} ... t_{n}, [/s]\}$ where $[/s]$ is a special token. Then the Transformer encoder utilizes the embedding representation of input sequence, which is calculated as the adding of language embedding, position embedding and token embedding of corresponding tokens, to perform attention-style fusion and self-attention computation. As a result, the encoder will outputs a cross-lingual context representation matrix $H \in \mathbb{R}^{(m+n+3) \times d}$, where $d$ is the vector dimension of the last layer of the encoder. In this way, the model can attend to both source-side terminology and target-side sentence, encouraging the model to learn and align the source and target representations.

\textbf{Span Detector:} Given the representation matrix output $H$ from Extractor\_Attn/Extractor\_Concat, we then input it to the linear layer, so as to separately predict the start index and the end index of the target terminology in target sentence. It can be formulated as follows:
\begin{equation}
    \textbf{p}_{start} = softmax(\textbf{W}_{start}\cdot H)
    \label{equ:p_start}
\end{equation}
\begin{equation}
    \textbf{p}_{end} = softmax(\textbf{W}_{end}\cdot H)
    \label{equ:p_end}
\end{equation}
where both of the $\textbf{W}_{start} \in \mathbb{R}^{d \times 2}$ and $\textbf{W}_{end} \in \mathbb{R}^{d \times 2}$ are linear layers with learnable parameters.

\textbf{Loss Function:} During the training, we separately calculate the loss of predicting the start index and end index of the target terminology, which are given as follows:
\begin{equation}
    \mathcal{L}_{start} = \textbf{CE}(\textbf{p}_{start},\textbf{y}_{start})
    \label{equ:loss_start}
\end{equation}
\begin{equation}
    \mathcal{L}_{end} = \textbf{CE}(\textbf{p}_{end},\textbf{y}_{end})
    \label{equ:loss_start}
\end{equation}
where \textbf{CE(*)} refers to cross-entropy computation. Then, the overall training objective to be minimized is as follows:
\begin{equation}
    \mathcal{L} = \frac{1}{2} (\mathcal{L}_{start} + \mathcal{L}_{end})
    \label{equ:loss_start}
\end{equation}
The two losses are jointly trained, with parameters shared at the linear layer. 

Note that in the \textbf{inference} phase, the start and end indexes will be predicted respectively. If both of them equal $0$ or the start index is larger than the end index, it means that there are no corresponding target terminology in the current sentence. If not, leading to the final extracted results of target terminology.

\section{Experiments}
We conduct experiments on Chinese$\to$English and English$\to$French 
e-commercial corpus to demonstrate the effectiveness of our proposed solutions.

\subsection{Data Construction}
\label{data-construction}

In this section, we describe the process of constructing e-commercial comparable corpus in detail.
We acquire English, Chinese, and French monolingual texts from the popular e-commerce platforms covering three product categories: clothes, toys, and outdoors. Monolingual texts under the same product category in different languages can be seen as e-commercial comparable corpus, which is not parallel sentence pairs but may contain potential parallel terminology pairs.

For each product category in all language pairs, we select frequent e-commercial terminologies from the monolingual sentences in source language (i.e., Chinese and English), and manually translate them into target language (i.e., English and French), which constitutes bilingual terminologies. In addition, we retrieve the monolingual sentences in target language containing the target terminology. 
If the target sentence contains the target terminology, we can construct a data pair of source terminology, target terminology and target sentence, which can be noted as the positive case, if not, negative case instead.  



\subsection{Experimental Setup}

\paragraph{Data Sets}
Following the data construction method described in Section \ref{data-construction}, we construct data of 
e-commercial bilingual terminology pairs, and the positive and negative cases of \{\textit{source terminology, target terminology, target sentence}\} pairs. 
For positive cases, we get the start and end indices of the target terminology in corresponding target sentence. For negative cases, we set both the start and end indices as 0.
Sequentially, we divide these data pairs into training, validation and test sets. 
In training/validation/test sets, the ratio of positive and negative cases remains at 1:1. 
The statistics of data sets are summarized in Table \ref{data_training}.

For cross-lingual pre-training in e-commerce, we use all the available monolingual e-commercial title corpus to perform MLM, which contains 5.5M, 7.4M, 4.1M for English, Chinese and French, respectively. 
Specially, we conduct TLM over the bilingual terminology pairs and the positive portion of the training set. The training sets of bilingual terminology pairs consist of 21,500 for Chinese$\to$English and 19,500 for English$\to$French.

\begin{table}[]
\caption{Statistics of the date sets. }
\centering
\begin{tabular}{|l|l|ccc|}
\hline
\multicolumn{2}{|c|}{\multirow{2}{*}{Data Sets}}      & \multicolumn{3}{c|}{  E-commercial Categories  } \\ \cline{3-5} 
\multicolumn{2}{|c|}{}                                & clothes    & toys     & outdoors \\ \hline \hline
\multirow{3}{*}{zh$\to$en} & training set   & 0.68M      & 0.46M   & 0.60M    \\
                                     & validation set & 1000       & 1000    & 1000     \\
                                     & test set       & 2694       & 2426    & 2396     \\ \hline
\multirow{3}{*}{en$\to$fr} & training set   & 0.61M      & 0.61M   & 0.61M    \\
                                     & validation set & 1000       & 1000    & 1000     \\
                                     & test set       & 2266       & 2442    & 2334     \\ \hline
\end{tabular}
\label{data_training}
\end{table}

\begin{table*}[htbp]
\caption{Experimental results(\%). ``RAND", ``MLM$_{\rm{eco}}$" and ``TLM$_{\rm{eco}}$" denote that our Extractor\_Attn and Extractor\_Concat models are initialized by random, MLM and TLM trained in e-commerce, respectively. ``avg." refers to the average performance of all three categories in Chinese$\rightarrow$English or English$\rightarrow$French.}
\centering
\begin{tabular}{llcccccccc}
\bottomrule[1.2pt]
\multicolumn{2}{l}{\multirow{2}{*}{System}}         & \multicolumn{4}{c}{zh$\rightarrow$en} & \multicolumn{4}{c}{en$\rightarrow$fr} \\
\cmidrule(lr){3-6} \cmidrule(lr){7-10}
\multicolumn{2}{c}{}                                & clothes  & toys     & outdoors& avg.    & clothes  & toys     & outdoors& avg.    \\
\bottomrule[1.2pt]
\multirow{4}{*}{Baselines}         & NMT            & 51.51   & 44.42 & 43.32 & 46.42 & 31.60   & 28.09 & 28.00 & 29.23 \\
                                   & SMT            & 54.31   & 47.13 & 47.63 & 49.69 & 39.81   & 36.69 & 36.82 & 37.77 \\
                                   & Multiple MT Voting & 47.92   & 56.22 & 53.84 & 52.66 & 53.40   & 57.41 & 56.93 & 55.91 \\
                                   & Seq2Seq-Term   & 65.26   & 50.78 & 54.01 & 56.68 & 41.66   & 30.06 & 32.65 & 34.79 \\
                                   & CLSP$_{\rm{eco}}$\cite{nagata2020supervised} & 86.06 & 73.52 & 74.81 & 78.13 & 70.84 & 64.92 & 68.46 & 68.07 \\  \hline \hline
\multirow{3}{*}{Extractor\_Attn}   & RAND           & 77.13   & 56.55 & 57.51 & 63.73 & 53.84   & 41.69 & 43.19 & 46.24 \\
                                   & MLM$_{\rm{eco}}$      & 84.86   & 70.57 & 73.54 & 76.32 & 67.87   & 62.33 & 61.30 & 63.83 \\
                                   & TLM$_{\rm{eco}}$      & 85.67   & 74.11 & 75.38 & 78.39 & 70.52   & 66.50 & 70.35 & 69.12 \\ \hline
\multirow{3}{*}{Extractor\_Concat} & RAND           & 83.96   & 69.41 & 70.45 & 74.61 & 76.00   & 62.16 & 66.36 & 68.17 \\
                                   & MLM$_{\rm{eco}}$      & 92.92   & 87.07 & 86.48 & 89.59 & 90.86   & 88.92 & 87.05 & 88.94 \\
                                   & TLM$_{\rm{eco}}$     & \textbf{94.43}  & \textbf{91.51}  & \textbf{90.23}  & \textbf{92.06}  & \textbf{92.58} & \textbf{90.25}  & \textbf{90.54} & \textbf{91.12} \\

\bottomrule[1.2pt]
\end{tabular}
\label{tbl:mainResult}
\end{table*}

\paragraph{Training Configuration}
For cross-lingual pre-training stage, we conduct joint byte-pair encoding (BPE) on the monolingual or comparable corpora of both languages with a shared vocabulary. We use the cross-lingual pretrained models released by XLM\footnote{https://github.com/facebookresearch/XLM} for the model initialization. During pre-training, following Conneau and Lample\cite{conneau2019cross}, 15\% of BPE tokens are selected to be masked. Among the selected tokens, 80\% of them are replaced with [MASK] token, 10\% are replaced with a random BPE token within the vocabulary, and 10\% remain unchanged. 

For target terminology extraction phase, we adopt the commonly used Transformer encoder with 1024 embedding/hidden units, 4096 feed-forward filter size, 6 layers and 8 heads per layer as the basic. During training, the batch size is set to 128 and the sentence length is limited to 100 BPE tokens. We employ the Adam \cite{kingma2014adam} optimizer with $lr$ = 0.0001, $t_{warm\_up}$ = 4000 and $dropout$ = 0.1.

\paragraph{Evaluation Metric}
During evaluating, we calculate the accuracy whether the model correctly predict both the start and end indices of the target-side terminology as follows:
\begin{equation}
    accuracy = \frac{Nums_{correct}}{Nums_{all}}
    \label{equ:acc}
\end{equation},
where $Nums_{correct}$ denotes the number of cases correctly predicted by the model, and ${Nums_{all}}$ denotes the number of all cases in the test set.

\subsection{Baselines}
We adopt the following methods as our baselines:

\begin{itemize}

\item \textbf{NMT/SMT:} We take the task of bilingual terminology extraction as an MT problem, i.e., bilingual terminology generation. 
Source terminology is fed into the MT model and the output sequence is target terminology. 
We adopt Transformer\footnote{ https://github.com/pytorch/fairseq/tree/v0.9.0. We use $\rm Transformer_{base}$ as our model. }\cite{Vaswani2017AttentionIA} and Moses\footnote{ http://www.statmt.org/moses/. We use the default setting of Moses. } as the NMT model and SMT system respectively. We measure whether the model correctly generates the entire target terminology as equation \ref{equ:acc}.
\item \textbf{Multiple MT Voting:}
We firstly utilize \textit{Google}\footnote{https://translate.google.com/}, \textit{Baidu}\footnote{https://fanyi.baidu.com/}, \textit{Youdao}\footnote{http://fanyi.youdao.com/}, \textit{bing}\footnote{https://bing.com/translator} and \textit{sogou}\footnote{https://fanyi.sogou.com/text} Translate Systems to directly translate the source terminology in our test set and get the corresponding translation candidates. Then we vote according to the results of different MT systems, with the highest number of votes as the final translation. We measure whether the final translation is the correct target terminology as equation \ref{equ:acc}.

\item \textbf{Cross-Language Span Prediction in E-Commerce (CLSP$_{\rm{eco}}$) \cite{nagata2020supervised}:} Cross-language span prediction method has been used for neural word alignment\cite{nagata2020supervised}, which can also be applied in e-commercial bilingual terminology extraction. To adapt it to e-commerce domain, we fine-tune multilingual BERT with e-commercial monolingual corpora. Then we formalize the task as SQuAD-style span prediction problem and solve it with the fine-tuned multilingual BERT as they propose in the paper.

\item \textbf{Seq2Seq-Term:} We regard the task as a sequence-to-sequence (seq2seq) learning problem by encoding the input of source terminology concatenated with target sentence, and decoding the output sequence of target terminology. For positive cases, the model will decode the corresponding target terminology. While for negative cases, the model will decode a special token ``None'', which means the translation of the source terminology does not exist in target sentence. Our Seq2Seq-Term system builds on Transformer\footnote{https://github.com/pytorch/fairseq/tree/v0.9.0. We use $\rm Transformer_{base}$ as our model.}, a state-of-the-art seq2seq model, with the shared vocabulary between input and output. This baseline is most related to our approaches, since they utilize the same data resources.

\end{itemize}

\subsection{Main Results}
Table \ref{tbl:mainResult} presents the performance of our proposed approach and other baseline models on different categories of different language pairs. 
It is obvious that our approaches outperform the baselines significantly in all language pairs and categories, which strongly demonstrates the superiority of cross-lingual pre-training and our proposed bilingual terminology extraction models.

\vspace{5 pt}
\noindent  \textbf{Comparison between Baselines}
\vspace{5 pt}

\noindent The performances of SMT systems are consistently superior to NMT systems, which indicates that directly using SMT trained on bilingual terminology pairs is more suitable for the task of bilingual terminology generation than NMT. 
In particular, we can find that Multiple MT Voting achieves better performance, mainly because it acquires the final translation results by voting among several top-tier MT engines. Seq2Seq-Term performs best among all baselines in zh$\to$en, while worse than Multiple MT Voting and SMT in en$\to$fr.

\begin{figure*}[htbp]
    \centering
    \includegraphics[width=0.85\textwidth]{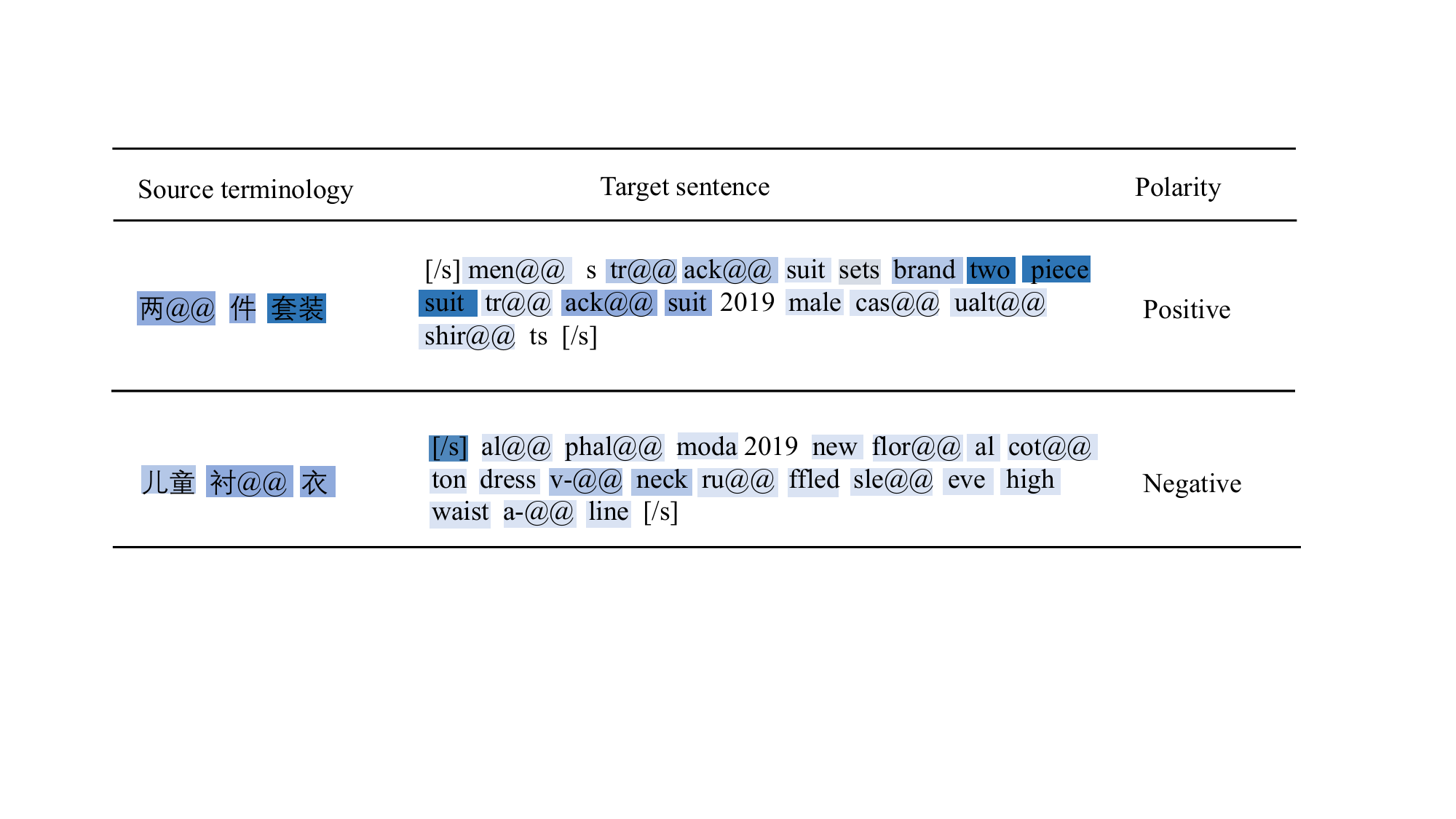}
    \caption{Visualized attention weights for source-side terminology and target-side sentence by Extractor\_Concat. ``Positive'' and ``Negative'' in column ``Polarity'' indicate whether the target sentence contains the corresponding translation of the source terminology or not.}
    \label{fig:attention}
\end{figure*}

\vspace{5 pt}
\noindent  \textbf{Comparison between Baselines and Our Proposed Approaches}
\vspace{5 pt}

\noindent Compared with various baselines, our proposed Extractor\_Attn and Extractor\_Concat with random initialization both show clear superiority, which demonstrates the effectiveness of our proposed bilingual terminology extraction models. Especially in comparison with the most related Seq2Seq-Term, our models show better performances, indicating that Extractor\_Attn and Extractor\_Concat are more suitable for the task of bilingual terminology extraction than the seq2seq method. 

\vspace{5 pt}
\noindent  \textbf{Comparison among Different Initialization Methods}
\vspace{5 pt}

\noindent When equipped with MLM$_{\rm{eco}}$ or TLM$_{\rm{eco}}$, our proposed Extractor\_Attn and Extractor\_Concat gain great improvements (+ 7.73\%-28.09\%), proving the significance of cross-lingual pre-training in e-commerce for the extraction models. Specially, models initialized with TLM$_{\rm{eco}}$ perform consistently better than those initialized with MLM$_{\rm{eco}}$ in all product categories. Obviously, the models could learn rich cross-lingual alignment information by TLM$_{\rm{eco}}$, which encourages the extraction models to better distinguish and even extract the target-side terminology.  

\vspace{5 pt}
\noindent  \textbf{Comparison between Extractor\_Attn and Extractor\_Concat}
\vspace{5 pt}

\noindent  Particularly, when comparing Extractor\_Attn and Extractor\_Concat, we note that Extractor\_Concat outperforms Extractor\_Attn under all model initialization conditions. Moreover, Extractor\_Concat initialized with TLM$_{\rm{eco}}$ obtains the best performances in all languages and all categories. It is because that Extractor\_Concat conducts self-attention computation on both source-side terminology and target-side sentence at the same time, while Extractor\_Attn calculates self-attention on source-side terminology and target-side sentence separately. We argue that Extractor\_Concat learns richer cross-lingual semantic relationship between source terminology and target sentence, and pay more attention to the most related span of target-side sentence.

\section{Analysis}

\begin{table}[t]
\caption{Performances(\%) of Extractor\_Concat with or without source-side terminologies, with different initialization parameters in Chinese$\rightarrow$English.}\label{tbl:w/o_src}
\centering
\begin{tabular}{ll|cccc}
\bottomrule[1.5pt]
                                                                  &      &clothes & toys     & outdoors & avg.    \\ \hline
\multicolumn{1}{l|}{\multirow{3}{*}{w/o source term}}             & RAND & 31.55  & 31.90 & 38.81 & 34.09 \\
\multicolumn{1}{l|}{}                                             & MLM$_{\rm{eco}}$  & 35.12  & 33.97 & 40.48 & 36.52 \\
\multicolumn{1}{l|}{}                                             & TLM$_{\rm{eco}}$  & 35.78  & 34.38 & 40.90 & 37.02 \\ \hline
\multicolumn{1}{l|}{\multirow{3}{*}{w/ source term}}              & RAND & 83.96  & 69.41 & 70.45 & 74.61 \\
\multicolumn{1}{l|}{}                                             & MLM$_{\rm{eco}}$  & 92.92  & 87.07 & 86.48 & 89.59 \\
\multicolumn{1}{l|}{}                                             & TLM$_{\rm{eco}}$  & 94.43  & 91.51 & 90.23 & 92.06 \\
\toprule[1.5pt]
\end{tabular}
\end{table}

\begin{table}[t]
\caption{Performances(\%) of Extractor\_Concat with or without the top self-attention layer, with different initialization parameters in Chinese$\rightarrow$English.}\label{tbl:w/o_attn}
\centering
\begin{tabular}{ll|cccc}
\bottomrule[1.5pt]
                                                                  &      &clothes & toys     & outdoors & avg.    \\ \hline
\multicolumn{1}{l|}{\multirow{3}{*}{w/ self-attn layer}}              & RAND & 83.96  & 69.41 & 70.45 & 74.61 \\
\multicolumn{1}{l|}{}                                             & MLM$_{\rm{eco}}$  & 92.92  & 87.07 & 86.48 & 89.59 \\
\multicolumn{1}{l|}{}                                             & TLM$_{\rm{eco}}$  & 94.43  & 91.51 & 90.23 & 92.06 \\ \hline
\multicolumn{1}{l|}{\multirow{3}{*}{w/o self-attn layer}}             & RAND & 83.67  & 68.26 & 69.78 & 73.90 \\
\multicolumn{1}{l|}{}                                             & MLM$_{\rm{eco}}$  & 92.72  & 86.69 & 85.64 & 88.35 \\
\multicolumn{1}{l|}{}                                             & TLM$_{\rm{eco}}$  & 94.21  & 91.18 & 90.12 & 91.84 \\

\toprule[1.5pt]
\end{tabular}
\end{table}

\subsection{Effect of Source Terminology}

\noindent In our proposed Extractor\_Concat, source terminology and target sentence are concatenated as an input sequence to the model, forming the final representation after self-attention computation. We wonder whether the model really learns the semantic relationship between source terminology and target sentence, or just extracts the target terminology depending on target sentence by simple positional recognition. Therefore, we remove the source terminology and take only the target sentence as input, attempting to predict target terminology just dependent on target sentence. Table \ref{tbl:w/o_src} shows the performance of Extractor\_Concat with or without source-side terminologies in Chinese$\rightarrow$English. We can observe that without source-side terminologies, the performances drop notably, which demonstrates the effect of interactiveness between source-side terminologies and target-side sentences.

\subsection{Effect of Last Self-attention Layer}
In our proposed Extractor\_Concat, self-attention layer is employed on the encoder output to obtain the cross-lingual context representation. We do ablation study to show the contribution of the top self-attention layer. Table \ref{tbl:w/o_attn} the performance of Extractor\_Concat with or without the top self-attention layer in Chinese$\rightarrow$English. It shows that the performance of Extractor\_Concat decreases when the top self-attention layer is removed, which demonstrates the significance of self-attention layer.

\subsection{Contextualized Word Representation}

\noindent To further investigate the effects of cross-lingual alignment, we sample two pairs of source terminology and target sentence from the Chinese$\rightarrow$English validation sets, and visualize the attention weights on them in Figure \ref{fig:attention}. The color depth indicates the importance degree of the weight,
the darker the more important. As can be seen, the semantic similarity between source and target terminology are able to be captured. In the positive example, \begin{CJK}{UTF8}{gbsn}
\textit{两@@ 件 套装}
\end{CJK} matches \textit{two piece suit}, which are mutual translations. While in the negative example, \begin{CJK}{UTF8}{gbsn}
\textit{儿童 衬@@ 衣}
\end{CJK} matches \textit{[/s]} in the beginning of target sentence, since target sentence does not contain its translation in target language.

\subsection{Effect of Training Data Size}

\begin{figure}[t]
    \centering
    \includegraphics[width=0.48\textwidth]{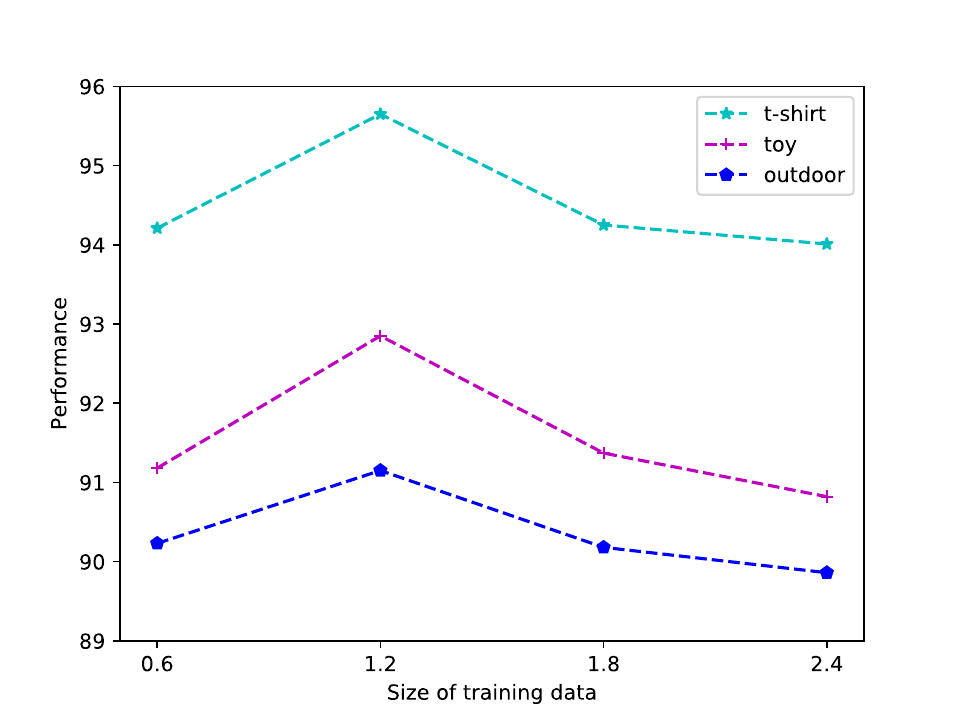}
    \caption{Performances(\%) of varying size(M) of training samples for  Extractor\_Concat  initialized by TLM$_{\rm{eco}}$.}
    \label{fig:size}
\end{figure}

\noindent We expand the number of positive and negative cases in the training sets (the ratio remains 1:1), so as to study the effect of the training data size on the performance. Table \ref{fig:size} reports the performances of our proposed Extractor\_Concat with double, triple and quadruple training data, i.e., 1.2M, 1.8M or 2.4M.
It shows that when we expand the size of training data to twice, the performances of all categories improve significantly. But when the size of training data is expanded to triple and quadruple, the performances of all categories drop sharply. More training data will not always lead to improvements of systems performances, suggesting that our model have already learnt enough cross-lingual semantic information with limited training data.
 
\section{Conclusion}
In this paper, we propose a new task of extracting bilingual terminologies from non-parallel comparable corpus in e-commerce and construct corresponding data sets. We apply a two-stage neural framework to tackle this task. When equipped with cross-lingual pre-training in e-commerce, our proposed Extractor\_Concat and Extractor\_Attn can extract the corresponding target terminology by fully utilizing the deep semantic relationship between source-side  terminology and target-side sentence. 
Experimental results show that our methods outperform all strong baselines in all categories on the Chinese$\rightarrow$English and English$\rightarrow$French language pairs.
As far as we know, we are the first to utilize cross-lingual pre-training and extraction model to solve the problem of extracting bilingual terminologies from non-parallel e-commerce corpora. We hope that our work will encourage the introduction of new paradigms for bilingual terminology extraction or other relevant research.

\section*{Acknowledgment}

The authors would like to thank the anonymous
reviewers for the helpful comments. This work was
supported by Project Funded by the Priority Academic Program Development of Jiangsu Higher Education Institutions, and was also partially supported by the joint research project of Alibaba and Soochow University.







\end{document}